\documentclass{article}

\usepackage{amsmath,epsfig}
\usepackage{amssymb}
\usepackage{microtype}
\usepackage{graphicx}
\usepackage{subfigure}
\usepackage{booktabs} 
\usepackage{hyperref}

\usepackage[accepted]{icml2019}

\icmltitlerunning{McMIL with Exact Likelihood}

\begin{document}

\twocolumn[
\icmltitle{A Multiclass Multiple Instance Learning Method with Exact Likelihood}
\icmlauthor{Xi-Lin Li, lixilinx@gmail.com}{}






\vskip 0.3in
]




\begin{abstract}
We study a multiclass multiple instance learning (MIL)  problem where the labels only suggest whether any instance of a class exists or does not exist in a training sample or example. No further information, e.g., the number of instances of each class,  relative locations or orders of all instances in a training sample, is exploited. Such a weak supervision learning problem can be exactly solved by maximizing the model likelihood fitting given observations, and finds applications to tasks like multiple object detection and localization for image understanding. We discuss its relationship to the classic classification problem, the traditional MIL, and connectionist temporal classification (CTC). We use image recognition as the example task to develop our method, although it is applicable to data with higher or lower dimensions without much modification. Experimental results show that our method can be used to learn all convolutional neural networks for solving real-world multiple object detection and localization tasks with weak annotations, e.g., transcribing house number sequences from the Google street view imagery dataset.   
\end{abstract}

\section{Introduction}

Traditionally, a classification or detection model is trained on data with detailed annotations showing the class label and location of each instance. Preparing such training data requires a lot of manual work. Segmentation of natural data like images and audios into regions or pieces just containing one instance of a class could be expensive, artificial and error prone due to the blurred boundaries among instances. Furthermore, models learned on such data typically expect the test data are preprocessed in similar ways. This significantly limits their applications. Let us take the image classification problem as an example task to make the above statements concrete. For example, both the classic convolutional neural network (CNN) LeNet-5 \cite{LeCun98} and moderns ones like the AlexNet \cite{alexnet} assume that one image only contains one instance of a class during both the train and test stages. Their last one or several layers are fully connected. Thus, they might further require the train and test images to have the same sizes. These limitations complicate their usages in reality where images come with diverse sizes, resolutions, and one image could be cluttered with multiple instances from one or multiple classes. More sophisticated object detection and localization models and methods, e.g., region CNN (R-CNN), single shot detectors (SSD), are proposed to address these challenges. A recent review of such methods is given in \cite{Huang17}. However, learning these advanced models requires training datasets with detailed annotations showing the class label and location of each object in all the training images.

Inspired by the traditional MIL method, this paper proposes an alternative solution to the classification or detection problem by relaxing the annotation requirements. For each class and each training sample, we only provide a binary label showing whether any instance of this class exists in this sample or not. As a result, the model has no need to predict details such as the number of instances,  their relative locations or orders, etc.. One might doubt the usefulness of models learned from such weak labels. However, the very simplicity of this setting might enable us to design general and elegant solutions to many tough problems. Figure 1 demonstrates the usage of an all convolutional network trained by our method on detecting the house numbers in images from the street view house numbers (SVHN) dataset \cite{svhn}. Without accessing to any detailed annotation such as the ground truth bounding boxes of digits in the images during the training stage, the learned all convolutional network can recognize and localize each digit in the images without further processing. Readers familiar with MIL may see the relationship between our method and MIL. However, their differences will be clear as the paper expands. Notably, our method uses binary {vector} label and exact bag probability for maximum likelihood model parameter estimation, while MIL typically assumes binary label and uses inexact bag probability or other loss functions for training, where a bag is a group of instances sharing a group level label. Our method can be used to learn the model from sketch, while task specific designs may play an important role in successfully applying the traditional MIL method. 

\begin{figure}
	\vskip 0.2in
	\begin{center}
	\includegraphics[width=\columnwidth]{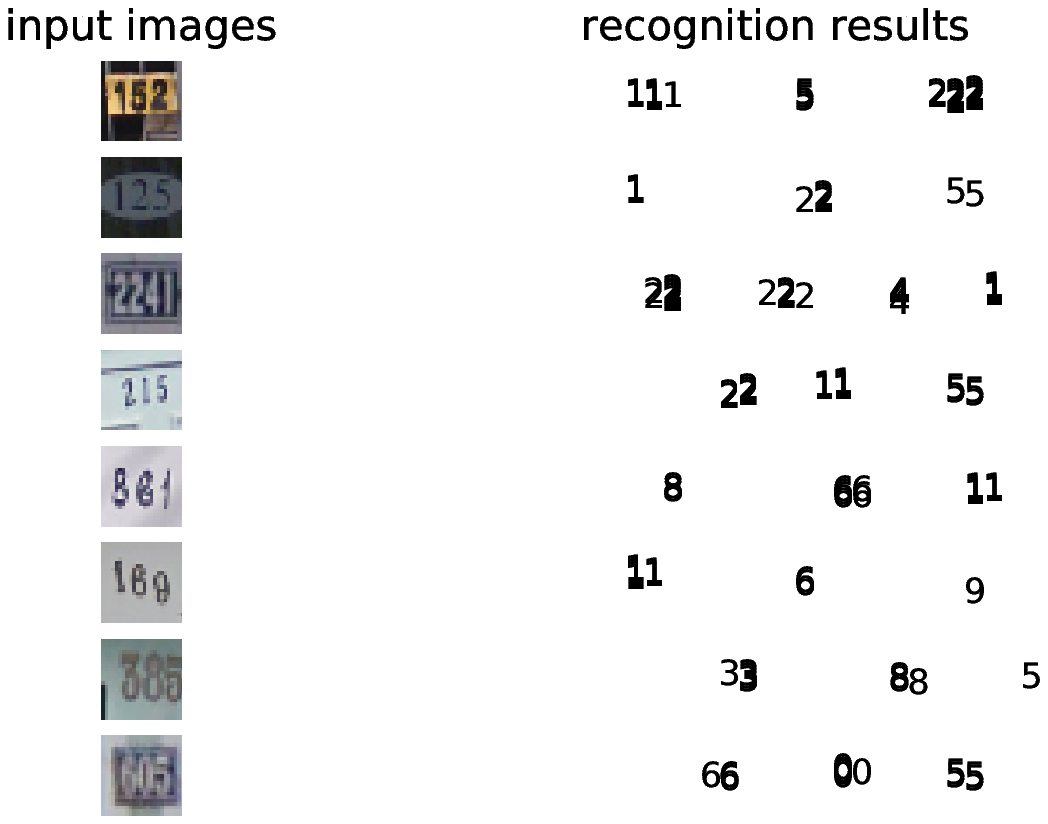}\\
	\caption{ Application of our method to the street view house number recognition task. On the left side are eight image crops with house numbers randomly selected from the test set, and on the right side are the detection results generated by an all convolutional network trained with our method. One can readily read the digit sequences by replacing successive and repetitive detected digits with a single and the same digit, i.e., a simple clustering process. For example, the top right sequence ``$\quad 11111 \quad  55 \quad 22222 \quad$'' is transcribed to ``$152$''.   }
\end{center}
\vskip -0.2in
\end{figure}

\section{Overview of our method}

We use image recognition as the example task to introduce our notations and formulation of the problem. However, one should  be aware that our method is not limited to image recognition with CNN. For example, one might use our method to train recurrent neural networks (RNN) for detecting certain audio events in an audio stream, or apply it to applications where the instances do not have any temporal or spatial orders, i.e., a bag of unsorted instances.  

\subsection{Binary vector label}

We consider an image classification or recognition problem. Let $\bold{I}$ be an input image tensor of size $(N, H, W)$, where $N$, $H$ and $W$ are the number of channels, height and width of image, respectively. The number of classes is $C$. Without loss of generality, we assume that these class labels are chosen from set $\mathbb{C}=\{1, 2, \ldots, C\}$. Each image $\bold{I}$ is assigned with a vector label $\pmb l = [l_1, l_2, \ldots, l_C]$ with length $C$, where $l_i$ can only take binary values, e.g., $0$ or $1$. By convention, $l_i=1$ suggests that at least one instance or object of the $i$th class exists in the associated image, and $l_i=0$ means that no instance of the $i$th class exists in the image. It is clear that label $\pmb l$ can take at most $2^C$ distinct values.  It is convenient to represent $\pmb l$ as a set denoted by $\mathbb{L}=\{ \ell_1, \ell_2, \cdots, \ell_L\}$, where $l_{i}=1$ if $i \in \mathbb{L}$, and $l_i=0$ when $i$ is not in set $\mathbb{L}$. Clearly, $\mathbb{L}$ only needs to contain distinct class labels. Unlike $\pmb l$, $\mathbb{L}$ may have variable length. Specifically, we have $\mathbb{L}=\{\}$, an empty set, when $\pmb l$ is a vector of zeros, and $\mathbb{L}=\mathbb{C}$ when $\pmb l$ is a vector of ones. We typically put $\bold{I}$ and its associated label $\pmb l$ or  $\mathbb{L}$ together as pair $(\bold{I}, \pmb l)$ or $(\bold{I}, \mathbb{L})$. One should not confuse the class label $\ell$ for an instance of the image and the label $\pmb l$ or  $\mathbb{L}$ for the whole image.  

A few examples will clarify our conventions. We consider the house number recognition task demonstrated in Figure 1. There are ten classes, and we assign class labels $1,2,\ldots,9,10$ to digits $1,2,\ldots, 9,0$, respectively. The third input image contains four digits, $2,2,4$ and $1$. Thus, its binary vector label will be $\pmb l=[1,1,0,1,0,0,0,0,0,0]$. Equivalently, we can set its label to $\mathbb{L}=\{1,2,4\}$ or $\{4,1,2\}$. When an image contains no digit, its label is either $\pmb l=[0,0,0,0,0,0,0,0,0,0]$, or simply $\mathbb{L}=\{\}$. We mainly adopt the notation of $\mathbb{L}$ in the rest of this paper.  

\subsection{Maximum likelihood model estimation}

Let $\mathcal{M(\pmb\theta)}$ be an image recognition model parameterized with trainable parameter vector $\mathcal{\pmb \theta}$. It accepts $\bold{I}$ as its input, and produces an output tensor $\bold{P}$ with shape $(C+1, M, N)$, where $M$ and $N$ are two positive integers determined by $\mathcal{M}$ and its input image sizes. Clearly, $\bold{P}$ is a function of $\pmb\theta$ and $\bold{I}$, although we do not explicitly show this dependence to simplify our notations. We define $\bold{P}$ as a probability tensor such that its $(\ell, m, n)$th element has meaning
\begin{equation}
p_{\ell,m,n}= \begin{cases}  {\rm Prb}({\rm emitting\, class \,label \,} \ell {\rm \, at \, position\, } (m,n) ), &\\
   \qquad\qquad\qquad\qquad\qquad\qquad\quad {\rm if\, } 1\le \ell\le C; & \\ 
 {\rm Prb}({\rm emitting\, class \,label \,} \phi {\rm \, at \, position\, } (m,n) ), &\\
   \qquad\qquad\qquad\qquad\qquad\qquad\quad {\rm if\, } \ell=C+1 & \\  \end{cases}
\end{equation}   
where ${\rm Prb}$ denotes the probability of an event, and $\phi$ can be understood as a class label reserved for the background. It is convenient to assign value $C+1$ to class label $\phi$ to simplify our notations. Thus, we will use $\phi$ and $C+1$ interchangeably. 
By definition, $\bold{P}$ is a nonnegative tensor, and has property 
\begin{equation}\label{normalized_P}
\sum_{\ell=1}^{C+1} p_{\ell,m,n} = 1,  \, {\rm for\, all\,} 1\le m\le M, 1\le n\le N
\end{equation}
For example, $\mathcal{M}(\pmb\theta)$ could be a CNN cascaded with a softmax layer to generate normalized probabilities at all locations. Then, we are possible to calculate the conditional probability for a label $\mathbb{L}$ given model $\mathcal{M}(\pmb \theta)$ and input image $\bold{I}$. We denote this probability with ${\rm Prb}(\mathbb{L}|\mathcal{M}(\pmb\theta), \bold{I})$. Now, the maximum likelihood model parameter vector is given by  
\begin{equation}\label{obj}
\pmb\theta^{\rm ml} = \arg\max_{\pmb\theta} E_{(\bold{I}, \,\mathbb{L})}\left[\log {\rm Prb}(\mathbb{L}|\mathcal{M}(\pmb\theta), \bold{I}) \right]
\end{equation}
where $E_{(\bold{I}, \,\mathbb{L})}$ denotes taking expectation over independent pairs of images and their associated labels. 
To calculate ${\rm Prb}(\mathbb{L}|\mathcal{M}(\pmb\theta), \bold{I})$ analytically, we do need to make the following working assumption. 

\emph{Assumption 1}: The model outputs at different locations are conditionally independent,
given the internal state of the network. 


Note that this is quite a standard assumption in solving similar problems with similar settings, e.g., hidden Markov model (HMM) and connectionist temporal classification (CTC) \cite{ctc}. All such models assume that the outputs at different locations are conditionally independent, given their internal states. This assumption is conveniently ensured by requiring that no feedback connections exist from the output layer to itself or the network.  

Let us consider two simple examples for the calculation of ${\rm Prb}(\mathbb{L}|\mathcal{M}(\pmb\theta), \bold{I})$. In the first example, we have ${\rm Prb}(\mathbb{L}|\mathcal{M}(\pmb\theta), \bold{I})=0$ when $|\mathbb{L}|> MN$ since the minimum number of instances already exceeds the total number of locations emitting class labels, where $|\mathbb{L}|$ is the order of $\mathbb{L}$. In the second example, we assume that  $\mathbb{L}=\{\ell\}$ and $\bold{P}$ has shape $(C+1, 2, 2)$. Recall that $\mathbb{L}=\{\ell\}$ suggests that at least one instance of the $\ell$th class appears in $\bold{I}$, and no instance of any other class shows up. Hence, only the following $15$ combinations of class labels emitted by the model are acceptable,     
\[ \begin{array}{c|c}
  \ell & \phi \\ 
  \hline
  \phi & \phi
 \end{array},\; 
\begin{array}{c|c}
  \phi & \ell \\ 
  \hline
  \phi & \phi
 \end{array},\; \begin{array}{c|c}
  \phi & \phi \\ 
  \hline
  \ell & \phi
 \end{array},\; \begin{array}{c|c}
  \phi & \phi \\ 
  \hline
  \phi & \ell
 \end{array},\; \begin{array}{c|c}
  \ell & \ell \\ 
  \hline
  \phi & \phi
 \end{array}  
\]
\[ \begin{array}{c|c}
  \ell & \phi \\ 
  \hline
  \ell & \phi
 \end{array},\; \begin{array}{c|c}
  \ell & \phi \\ 
  \hline
  \phi & \ell
 \end{array},\; \begin{array}{c|c}
  \phi & \ell \\ 
  \hline
  \ell & \phi
 \end{array},\; \begin{array}{c|c}
  \phi & \ell \\ 
  \hline
  \phi & \ell
 \end{array},\; \begin{array}{c|c}
  \phi & \phi \\ 
  \hline
  \ell & \ell
 \end{array}  
\]
\[ \begin{array}{c|c}
  \ell & \ell \\ 
  \hline
  \ell & \phi
 \end{array},\; \begin{array}{c|c}
  \ell & \ell \\ 
  \hline
  \phi & \ell
 \end{array},\; \begin{array}{c|c}
  \ell & \phi \\ 
  \hline
  \ell & \ell
 \end{array},\; \begin{array}{c|c}
  \phi & \ell \\ 
  \hline
  \ell & \ell
 \end{array},\; \begin{array}{c|c}
  \ell & \ell \\ 
  \hline
  \ell & \ell
 \end{array}  
\]
By Assumption 1, the probability of each combination is the product of the probabilities emitted at all locations. ${\rm Prb}(\mathbb{L}|\mathcal{M}(\pmb\theta), \bold{I})$ will be the sum of the probabilities of all feasible combinations, i.e., 
\begin{align*}
 &{\rm Prb}(\mathbb{L}|\mathcal{M}(\pmb\theta), \bold{I})= p_{\ell, 1, 1}p_{\phi, 1,2} p_{\phi, 2,1} p_{\phi,2,2}  \\
& \quad + p_{\phi, 1, 1}p_{\ell, 1,2} p_{\phi, 2,1} p_{\phi,2,2} + p_{\phi, 1, 1}p_{\phi, 1,2} p_{\ell, 2,1} p_{\phi,2,2} \\
& \quad + p_{\phi, 1, 1}p_{\phi, 1,2} p_{\phi, 2,1} p_{\ell,2,2} + p_{\ell, 1, 1}p_{\ell, 1,2} p_{\phi, 2,1} p_{\phi,2,2} \\
& \quad + p_{\ell, 1, 1}p_{\phi, 1,2} p_{\ell, 2,1} p_{\phi,2,2} + p_{\ell, 1, 1}p_{\phi, 1,2} p_{\phi, 2,1} p_{\ell,2,2} \\
& \quad + p_{\phi, 1, 1}p_{\ell, 1,2} p_{\ell, 2,1} p_{\phi,2,2} + p_{\phi, 1, 1}p_{\ell, 1,2} p_{\phi, 2,1} p_{\ell,2,2} \\
& \quad + p_{\phi, 1, 1}p_{\phi, 1,2} p_{\ell, 2,1} p_{\ell,2,2} + p_{\ell, 1, 1}p_{\ell, 1,2} p_{\ell, 2,1} p_{\phi,2,2} \\
& \quad + p_{\ell, 1, 1}p_{\ell, 1,2} p_{\phi, 2,1} p_{\ell,2,2} + p_{\ell, 1, 1}p_{\phi, 1,2} p_{\ell, 2,1} p_{\ell,2,2} \\
& \quad + p_{\phi, 1, 1}p_{\ell, 1,2} p_{\ell, 2,1} p_{\ell,2,2} + p_{\ell, 1, 1}p_{\ell, 1,2} p_{\ell, 2,1} p_{\ell,2,2} 
\end{align*}
The above naive method for calculating ${\rm Prb}(\mathbb{L}|\mathcal{M}(\pmb\theta), \bold{I})$ has complexity $\mathcal{O}(2^{MN})$ even in the case of $|\mathbb{L}|=1$. We provide two affordable methods for calculating ${\rm Prb}(\mathbb{L}|\mathcal{M}(\pmb\theta), \bold{I})$ in the next section. 

\section{Model likelihood calculation}

\subsection{Alternating series expression}

To begin with, let us introduce a quantity $\alpha_{\ell_1, \ell_2, \ldots, \ell_L}$. It denotes the probability of the event that at least one instance of any class from set $\{\ell_1, \ell_2, \ldots, \ell_L\}$ shows up in $\bold{I}$, and no instance from any other class shows up. The class label in set $\{\ell_1, \ell_2, \ldots, \ell_L\}$ can be $\phi$, i.e., $C+1$ by our convention. Still, all these class labels are required to be distinct. Thus, we have $L\le C+1$. With this notation, $\alpha_{\ell_1, \ell_2, \ldots, \ell_L}$ is given by
\begin{equation}\label{alpha}
\alpha_{\ell_1, \ell_2, \ldots, \ell_L} = \prod_{m=1}^M \prod_{n=1}^N \sum_{i=1}^L p_{\ell_i,  m, n}
\end{equation}
Specifically, we have 
\begin{equation}
\alpha_{\ell} = \prod_{m=1}^M \prod_{n=1}^N  p_{\ell,  m, n}, \quad \alpha_{1,2,\ldots, C+1}=1
\end{equation}

The following statement can be used to calculate ${\rm Prb}(\mathbb{L}|\mathcal{M}(\pmb\theta), \bold{I})$ efficiently. With a slight abuse of notations, we use ${\rm Prb}(\mathbb{L}|\bold{P})$ to denote this probability since $\bold{P}$ is uniquely determined by $\mathcal{M}(\pmb\theta)$ and $\bold{I}$. 

\emph{Proposition 1}: Let $\mathbb{L}=\{\ell_1, \ell_2, \ldots, \ell_L \}$ be a set of distinct class labels excluding $\phi$. The probability ${\rm Prb}(\mathbb{L}|  \bold{P})$ is given by the sum of the following $L+1$ terms
\begin{align}\nonumber
{\rm Prb}(\mathbb{L}|\bold{P}) = & (-1)^0 \alpha_{\ell_1, \ell_2, \ldots, \ell_L, \phi}   \\ \nonumber
& + (-1)^{1} \sum_{i=1}^L \alpha_{\ell_1, \ldots, \ell_{i-1}, \ell_{i+1}, \ldots, \ell_L, \phi} \\ \nonumber
& + \ldots   \\ \nonumber
& + (-1)^{L-3}\sum_{i=1}^L\sum_{j=i+1}^L\sum_{k=j+1}^L  \alpha_{\ell_i, \ell_j, \ell_k, \phi} \\ \nonumber
& +  (-1)^{L-2}\sum_{i=1}^L\sum_{j=i+1}^L  \alpha_{\ell_i, \ell_j, \phi} \\ \nonumber
 & +  (-1)^{L-1}\sum_{i=1}^L \alpha_{\ell_i, \phi} \\ \label{prop1}
& +  (-1)^L \alpha_{\phi}
\end{align}

We have outlined its proof in Appendix A. Here, we consider a few special cases to have an intuitive understanding of Proposition 1. By definition, we have 
$${\rm Prb}(\{\}|\bold{P}) =  \alpha_{\phi} =  \prod_{m=1}^M \prod_{n=1}^N  p_{C+1,  m, n} $$ 
When only instances of the $\ell$th class can show up, we have 
\begin{equation}\label{prb_1label}
{\rm Prb}(\{\ell \}| \bold{P}) =  \alpha_{\ell, \phi} -\alpha_{\phi}
\end{equation}
where the term $-\alpha_{\phi}$ compensates the probability of the event that only class label $\phi$ is emitted at all locations. With order $|\mathbb{L}|=2$, we have
\begin{equation}\label{P_l1_l2}
{\rm Prb}(\{\ell_1, \ell_2 \}|\bold{P}) =  \alpha_{\ell_1, \ell_2, \phi} -\alpha_{\ell_1, \phi} - \alpha_{\ell_2, \phi}   + \alpha_{\phi} 
\end{equation}
where the term $-\alpha_{\ell_1, \phi} - \alpha_{\ell_2, \phi} + \alpha_{\phi}$ compensates the probability that instance from either one of the two classes, $\ell_1$ and $\ell_2$, is missing, and $\alpha_{\phi}$ is here since it is counted twice in sum $\alpha_{\ell_1, \phi} + \alpha_{\ell_2, \phi}$. In general, we will observe the pattern of alternating signs in (\ref{prop1}). Note that Proposition 1 still holds when $|\mathbb{L}|>MN$. For example, when $M=N=1$, explicitly expanding (\ref{P_l1_l2}) with (\ref{alpha}) leads to
\begin{align*}
& {\rm Prb}(\{\ell_1, \ell_2 \}|\bold{P})  =(p_{\ell_1,1,1} + p_{\ell_2,1,1} + p_{\phi, 1, 1}) \\
& \qquad -(p_{\ell_1,1,1} + p_{\phi, 1, 1})-( p_{\ell_2,1,1} + p_{\phi, 1, 1}) + p_{\phi, 1, 1} = 0 
\end{align*}
which makes perfect sense since one location cannot emit two different class labels. 

\subsection{Recursive expression}

We introduce another quantity $\beta_{\ell_1, \ell_2, \ldots, \ell_L}$. It denotes the probability of the event that at least one instance of each class from set $\{\ell_1, \ell_2, \ldots, \ell_L \}$ appears in $\bold{I}$, and no instance of any other class shows up. Again, the class label in set $\{\ell_1, \ell_2, \ldots, \ell_L\}$ can be $\phi$, i.e., $C+1$ by our convention. All these class labels are distinct. Thus, we have $L\le C+1$ as well. By this definition, we have
\begin{equation}\label{one_class}
\beta_{\ell}=\alpha_{\ell} = \prod_{m=1}^M \prod_{n=1}^N p_{\ell, m,n}, \quad 1\le \ell \le C+1
\end{equation}
However, for $L>1$, we can only calculate $\beta$ recursively by relationship 
\begin{align}\nonumber 
\beta_{\ell_1, \ldots, \ell_L} = & \alpha_{\ell_1, \ldots, \ell_L} -\sum_{i=1}^L \beta_{\ell_i} -\sum_{i=1}^L\sum_{j=i+1}^L \beta_{ \ell_i, \ell_j}  \\  \label{beta_recursive}
& - \ldots - \sum_{i=1}^L \beta_{ \ell_1, \ldots, \ell_{i-1}, \ell_{i+1}, \ldots, \ell_L}  
\end{align}
where the initial conditions are given by (\ref{one_class}). The meaning of (\ref{beta_recursive}) is clear. On its right side, each $\beta$ term denotes the probability of an event that instances of certain classes from set $\{\ell_1, \ell_2, \ldots, \ell_L\}$ are missing. Hence, the left side of (\ref{beta_recursive}) gives the probability of the event that at least one instance of each class in set $\{\ell_1, \ell_2, \ldots, \ell_L\}$ shows up. 
With the above notations, ${\rm Prb}(\{\ell_1, \ldots, \ell_L\}|\bold{P})$ can be calculated using the following expression,
\begin{align}\nonumber 
& {\rm Prb}(\{\ell_1, \ldots, \ell_L\}|\bold{P}) = \\ \nonumber 
&\qquad  \alpha_{\ell_1, \ldots, \ell_L, \phi} -\beta_{\phi} - \sum_{i=1}^L (  \beta_{\ell_i} + \beta_{\ell_i, \phi}) \\  \nonumber 
& \qquad -\sum_{i=1}^L\sum_{j=i+1}^L ( \beta_{ \ell_i, \ell_j} + \beta_{ \ell_i, \ell_j, \phi}) - \ldots \\ \label{prb_in_beta}
& \qquad - \sum_{i=1}^L(  \beta_{ \ell_1, \ldots, \ell_{i-1}, \ell_{i+1}, \ldots, \ell_L}  + \beta_{ \ell_1, \ldots, \ell_{i-1}, \ell_{i+1}, \ldots, \ell_L, \phi}) 
\end{align}
Eq. $(\ref{prb_in_beta})$ provides the starting point for the proof of Proposition 1. 

Note that both the expressions given in (\ref{prop1}) and $(\ref{prb_in_beta})$ have complexity $\mathcal{O}(2^L\,MN)$. Neither  will be affordable for large enough $L$ \footnote{With settings $M=N=C=100$, $L>16$ is considered to be large as it takes seconds to evaluate (\ref{prop1}) for $L=16$ on our computer.}. The series in (\ref{prop1}) has alternating signs, and generally, its truncated version gives neither an upper bound nor a lower bound of ${\rm Prb}(\{\ell_1, \ldots, \ell_L\}|\bold{P}) $. Compared with (\ref{prop1}), one advantage of (\ref{prb_in_beta}) is that it allows us to truncate the series in (\ref{prb_in_beta}) to obtain an upper bound of ${\rm Prb}(\{\ell_1, \ldots, \ell_L\}|\bold{P}) $ since all $\beta$'s are nonnegative. A truncated version of the series given in $(\ref{prb_in_beta})$ could provide an affordable approximation for ${\rm Prb}(\{\ell_1, \ldots, \ell_L\}|\bold{P}) $ for arbitrarily large $L$. Fortunately, in practice, $L$ is seldom too large to make the expressions given in (\ref{prop1}) unaffordable. 

\subsection{Calculation with floating point arithmetic}

Generally, neither (\ref{prop1}) nor (\ref{prb_in_beta}) can be directly used to calculate ${\rm Prb}(\{\ell_1, \ldots, \ell_L\}|\bold{P})$ with floating point arithmetic due to the high possibility of underflow when calculating these $\alpha$'s or $\beta$'s. We always use their logarithms for the calculations. For example, let $x>y>0$ be two  positive numbers. The relationship   
\[ \log(x\pm y) = \log x + \log(1 \pm \exp(\log y - \log x)) \]
is proposed to calculate $x+y$ or $x-y$ when $x$ and $y$ are tiny. More details are given in our implementations. 

\section{Relationship to existing techniques}

\subsection{Relationship to the classic classification problem}

It is worthwhile to consider the classic classification problem where each sample only contains one instance of a class. For example, the commonly used image classification datasets, e.g., MNIST \cite{LeCun98} and ImageNet \cite{imagenet_cvpr09}, all assume the one-instance-per-image setting. The image classification task itself is of great interest, and it also is one of the most important building blocks in achieving more complicated tasks, e.g., object detection and localization. In our notations, we have $|\mathbb{L}|=1$ for this setting. Still, we do not assume the number of instances of the selected class, which may be preferred in situations where the instances are too crowded to count for their exact number. Let us consider the train and test stages with our method.          

During the train stage, our method maximizes the logarithm likelihood objective function in (\ref{obj}), where the probability ${\rm Prb}(\{\ell \}|  \bold{P})$ is given by (\ref{prb_1label}). When the probability tensor $\bold{P}$ has shape $(C+1, 1, 1)$, (\ref{prb_1label}) reduces to
\begin{equation}
{\rm Prb}(\{\ell \}|\bold{P}) = (p_{\ell, 1, 1} + p_{\phi, 1, 1} )- p_{\phi, 1, 1}=p_{\ell, 1, 1}
\end{equation}
Then, maximizing the logarithm likelihood objective function in (\ref{obj}) is equivalent to minimizing the cross entropy loss, a routine practice in solving the classic classification problem. 

During the test stage, we calculate the tensor $\bold{P}$ for a test sample, and scan ${\rm Prb}(\{\ell \}|\bold{P})$ for all $1\le \ell \le C$. The estimated class label is the one that maximizes ${\rm Prb}(\{\ell \}|\bold{P})$, i.e.,
\[ \hat{\ell} = \arg\max_{\ell} {\rm Prb}(\{\ell \}|\bold{P}) ,\qquad 1\le \ell\le C \]
Noting that ${\rm Prb}(\{\ell \}|\bold{P})=\alpha_{\ell, \phi} -\alpha_{\phi}$, and $\alpha_{\phi}$ is independent of $\ell$, we simply have 
\begin{equation}\label{1l_per_smpl_detection}
\hat{\ell} = \arg\max_{\ell}   \alpha_{\ell, \phi} ,\qquad 1\le \ell\le C
\end{equation}
For the classic classification problem, we may have no need to consider the class label $\phi$ since as the training samples, each test sample is supposed to be associated with a unique class label in range $[1,C]$. However, in many more general settings, e.g., multiple object detections, class label $\phi$ plays an important role in separating one instance of a class from the background and other instances, either belonging to the same or different classes.      

For a well trained classifier, the model typically tends to emit either class label $\phi$ or $\hat{\ell}$ at any location $(m, n)$. Thus, we should have $p_{\hat{\ell},  m, n} + p_{\phi,  m, n}\approx 1$ for any pair $(m, n)$. With Taylor series 
$$\log(z) = (z-1) - (z-1)^2/2 + \ldots$$
we have 
\begin{align*}
\log \alpha_{\hat{\ell}, \phi} =  & \sum_{m=1}^M \sum_{n=1}^N \log ( p_{\hat{\ell},  m, n} + p_{\phi,  m, n}) \\
\approx & \sum_{m=1}^M \sum_{n=1}^N ( p_{\hat{\ell},  m, n} + p_{\phi,  m, n} - 1)
\end{align*}
Thus, we could estimate the class label simply by 
\begin{equation}\label{1l_per_smpl_detection2}
\hat{\ell} \approx \arg\max_{\ell}  \frac{1}{MN}\sum_{m=1}^M \sum_{n=1}^N p_{{\ell},  m, n} ,\quad 1\le \ell\le C
\end{equation}
Our experiences suggest that (\ref{1l_per_smpl_detection}) and (\ref{1l_per_smpl_detection2}) give identical results most of the time. However, (\ref{1l_per_smpl_detection2}) is more intuitive, and closer to a CNN classifier with fully connected layers when one implements the average operation on the right side of (\ref{1l_per_smpl_detection2}) as an average pooling layer. 

\subsection{Relationship to weak supervision and MIL}

Considering that detailed data annotations are expensive, learning with weak supervision is promising, and becoming an active research direction recently. For example, \cite{Papadopoulos17} provides a mean to reduce the annotation work,  \cite{Blaschko10} proposes a framework to learn from both fully and weakly annotated data, and \cite{Gonzalez16} and \cite{Zhou18} give reviews and taxonomies of weakly supervised learning by considering the instance-label relationship and completeness/accurateness of supervision, respectively. Weak supervision and annotation are rather ambiguous terms. They could have quite different forms and meanings. Among them, our method is most closely related to MIL \cite{Carbonneau18, Zhang14} since it also assumes a binary label for each class. The work in \cite{Wu15} further combines deep learning and MIL, and applies MIL to the bag of crops generated by region proposals for image classification and auto-annotation. Still, our method is different from the traditional MIL in several ways. One important difference is that the standard MIL only considers binary classifications, while our method applies to both binary and multiclass classifications. Another important difference is that Proposition 1 gives the exact model likelihood, while MIL typically uses approximate bag probabilities, or more often, certain loss functions for classification. 

Let us consider a binary classification problem where $1$ and $\phi$ are the positive and negative labels, respectively. In our notations, probability of a negative image bag is straightforwardly shown to be $\alpha_{\phi}=\prod_{m=1}^M \prod_{n=1}^N p_{\phi, m,n}$. Many MIL methods simply take $\max_{m,n}p_{1, m, n}$ as the probability of a positive image bag. Although this approximation is inaccurate, it readily generalizes to multiclass MIL. For example, \cite{Pathak15} approximates the probability of a positive image bag with multiple class labels $\mathbb{L}=\{\ell_1, \ldots, \ell_L\}$ with $\prod_{\ell \in \mathbb{L}} \max_{m,n}p_{\ell, m, n} $ to arrive cost function $- \sum_{\ell \in \mathbb{L}} \left( \log \max_{m,n}p_{\ell, m, n} \right) /{|\mathbb{L}|}$ for model parameter optimization. One-vs.-rest is another strategy for solving the multiclass MIL via standard binary MIL. Although these solutions are shown to be successful for numerous applications, especially when the models are convex, we do not get reasonably good performance when tested on our multiple object detection tasks (Supplementary Material A). One possible reason may be the optimization difficulty since our models are highly non-convex and $\max$ operation is not differentiable. Our method also requires less domain specific knowledge, e.g., pre-training and region proposal, which are required by \cite{Wu15, Pathak15}. To our best knowledge, we are the first to derive the exact model likelihood for multiclass MIL under rather general settings.

\subsection{Relationship to connectionist temporal classification (CTC)}

CTC provides another training framework for learning without knowing the exact location of each instance \cite{ctc}. However, it does assume that all instances can be ordered in certain ways, e.g., temporal order, such that a dynamic programming like method can be used to calculate the model likelihood efficiently. Thus, it mainly applies to sequence processing tasks. Repetitive class labels in CTC are meaningful as they correspond to repetitive appearances of instances from the same class. For data with higher intrinsic dimensions, e.g., images with randomly scattered objects, we typically can only define a partial order, not a strict order, relationship for all those instances from a sample based on their relative positions. To our knowledge, there might not exist an efficient way to calculate a CTC like model likelihood given the complete partial orders of all instances. Hence, we take one step back and only assume the knowledge of the existence of instance of any class. This considerably reduces the annotation work, and also simplifies the model likelihood calculation.      

\section{Experimental results}

We only consider the image classification problems here, although our method could apply to other tasks, e.g., audio event detection. To our framework, the main difference among these tasks is that the probability tensor $\bold{P}$ will have different orders. Proposition 1 always holds regardless of the detailed forms of $\bold{P}$. In the following image recognition tasks, we only consider fully convolutional networks without pooling layer or short cut connection for design simplicity. Subsampling or decimation, if necessary, is achieved by convolutional layers with stride larger than $1$. A Newton type method \cite{anonymous2019learning} using normalized step size is adopted for model likelihood optimization to save tuning efforts. With $E$ epochs of iterations, we typically set the learning rate to $0.01$ for the first $0.5E$ epochs, and $0.001$ for the last $0.5E$ epochs, where $E$ takes values of tens. Batch sizes are $64$ for smaller models and $32$ for larger ones due to limited GPU memories. We only report the comparison results with strongly supervised learning methods using detailed annotations. The traditional MIL does not perform well without task specific designs like pre-training and region proposal, and we report its performance in Supplementary Material A for interested readers. We always train our models from sketch. Pytorch implementations reproducing the following reported results are available at \url{https://github.com/lixilinx/MCMIL}. Any detail not reported here can be found in our implementation package.  

\subsection{Application to the classic classification problems}

\subsubsection{MNIST handwritten digit recognition} 

We have tested a CNN model with five layers for feature extractions, and one last layer for detection. All the convolution filters have kernel size $5\times 5$. Decimations and zero paddings are set to let $\bold{P}$ has shape $(11, 4, 4)$. A baseline CNN model having almost identical structure is considered as well. We just apply less zero paddings in the baseline model to make its $\bold{P}$ has shape $(10, 1, 1)$ such that the traditional cross entropy loss can be used to train it. Both models have about $0.41$ M coefficients to learn. With these settings, test classification error rates of the baseline model and ours are $0.54\pm0.03\%$ and $0.48\pm 0.04\%$, respectively. Our method performs slightly better on this task.  

\subsubsection{CIFAR10 image classification}

We have tested a CNN model with nine layers for feature extractions, and one last layer for detection. All convolutional filters have kernel size $3\times 3$. Decimations and zero paddings are set to let $\bold{P}$ has shape $(11, 4, 4)$. A similar baseline model trained by minimizing cross entropy loss is considered as well. Both models have about $2.7$ M coefficients to learn. With these settings, test classification error rates of the baseline model and ours are $8.4\pm 0.4\%$ and $8.7\pm 0.3\%$, respectively. The baseline model performs slightly better here. Note that for this task, deeper and larger models can achieve better test error rate performances. The purpose of this experiment is not to compete with those state-of-the-art results, but to empirically show that replacing the traditional cross entropy loss with ours does not lead to meaningful performance loss. Actually, our performances are no worse than those of the all convolutional nets reported in \cite{Springenberg15}.  

\subsection{Extended MNIST experiment}

\begin{figure}
\vskip 0.2in
\begin{center}
	\includegraphics[width=0.9\columnwidth]{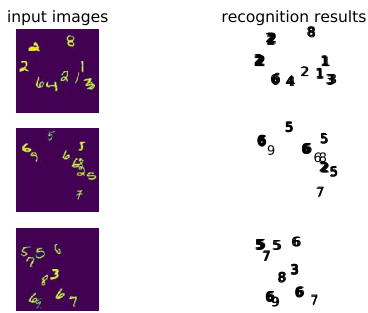}\\
	\caption{ Models trained in the extended MNIST experiment can recognize randomly scattered handwritten digits in images of arbitrary sizes  without any further processing.   }
\end{center}
\vskip -0.2in
\end{figure}

We use synthesized MNIST data to learn the same CNN model in Section 5.1.1. We randomly select two handwritten digit images and nest them into a larger one. Then, this larger image and its label, which only tells that what digits appear in the image and what do not, are fed into our model for training. As a result, the model never get a chance to see any individual digit. Still, the learned model can recognize multiple randomly scattered handwritten digits in new test images with arbitrary sizes without any further processing, as shown in Figure 2. Here, the class label $\phi$ plays an important role in separating one instance from another. We have tested the learned model on the same MNIST test dataset, and the test classification error rate is $0.37\pm0.04\%$. The lowest one among ten runs starting from random initial guesses is $0.32\%$. To our knowledge, these are among the best test error rates on the MNIST dataset ever achieved without using any regularization, or affine and elastic distortions for data augmentation. 

\subsection{SVHN experiment}

We consider the street view house number recognition task \cite{svhn} in settings as realistic as possible. The task is to transcribe an image with house numbers to a string of digits. The training images come with very different sizes and resolutions. To facilitate the training, we take a tight \emph{square} crop containing all the digits of an image, and rescale it to size $64\times 64$. During the training stage, the model only knows what digits appear in the crops, and what do not. Thus, only limited annotation information is used to train our models. Note that most other works solving this task exploit more annotations and prior knowledge, e.g., the complete digit sequences,  their locations,  the minimum and maximum sequence lengths, etc. \cite{Goodfellow14, JBa15}. Furthermore, they use tight rectangle crops, which causes aspect ratio loss after rescaling the crops to size $64\times 64$. 

\begin{figure}
\vskip 0.2in
\begin{center}
	\includegraphics[width=0.8\columnwidth]{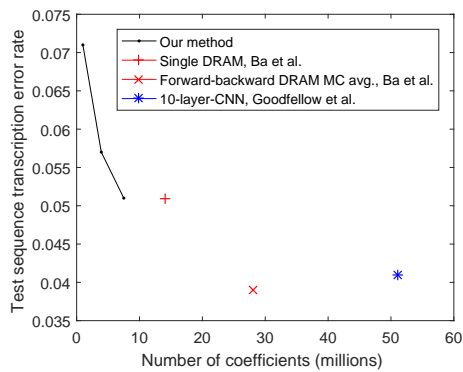}\\
	\caption{ Test sequence transcription error rate comparison among our method and strongly supervised ones (oracles) from \cite{Goodfellow14} and \cite{JBa15} on the SVHN dataset. DRAM and MC are the abbreviations of Deep Recurrent Attention Model and Monte Carlo, respectively. }
\end{center}
\vskip -0.2in
\end{figure}

We have trained three CNN models. All have $10$ layers, and consist of convolutional filters with kernel size $5\times 5$. Decimations and zero paddings are set to let $\bold P$ has shape $(11, 28, 28)$. As shown in Figure 1, our trained models are able to read the digits in the test crops. Still, we need to transcribe these readings into house number sequences. Currently, we use a very coarse rule-based transcription method to convert the recognized digits into a sequence. We only consider those horizontally oriented house numbers. As illustrated in Figure 1, we simply replace successive and repetitive detected digits with a single the same digit to obtain the transcriptions. However, this simple clustering method could yield incorrect transcriptions in several situations even when the model successfully recognizes all digits. Nevertheless, the overall performances of our models are still competitive. Figure 3 summarizes the test sequence transcription error rates of several compared methods. State-of-the-art test error rate for this task is about $3.9\%$ \cite{JBa15}, while our best one is about $5.1\%$. Although the performance gap is significant, our models use less coefficients, and are significantly simpler and more widely applicable. We have tried to further increase the model size, but the test performance gain is rather limited. The lack of an end-to-end, i.e., image-to-sequence, training cost might explain the performance gap between our method and the ones in \cite{Goodfellow14, JBa15}. 

\begin{figure}
\vskip 0.2in
\begin{center}
	\includegraphics[width=0.85\columnwidth]{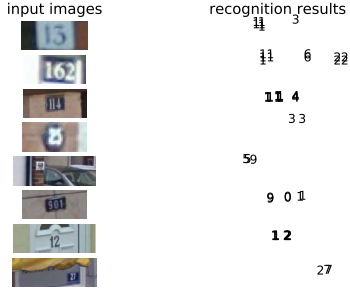}\\
	\caption{ Our models can successfully recognize house numbers in many of the original images without rescaling or using the ground truth bounding boxes. Still, they may fail when the original images have significantly higher resolutions or are cluttered with objects  resembling digits, e.g., letters and stripes, since our models are not exposed to such training data. }
\end{center}
\vskip -0.2in
\end{figure}

Salient advantages of our models are their simplicity and wider applicabilities. Figure 4 shows some examples where our models can recognize house numbers in the original images with different sizes and resolutions without rescaling or ground truth bounding boxes information. Clearly, our models are able to detect and locate each digit instance. A closer check suggests the following transcription error patterns. \emph{Clustering-error}: Ideally, one cluster should correspond to one instance. But, our transcription method is too coarse. It may fail to find the correct number of instances due to clustering error. \emph{Orientation-error}: Our transcription method assumes horizontally orientated house numbers. It fails when the house numbers are vertically orientated. \emph{Edge-vs.-1-error}: Many detection errors are due to detecting vertical edges as digit $1$, or failing to detect digit $1$, possibly regarding it as edges by the models. Indeed, it could be difficult to distinguish vertical edges and digit $1$ for our models since they are all locally connected convolutional networks without fully connected layers. Increasing the receptive field of the last detection layer may help to alleviate this issue. \emph{No-detection-error}: Our method may detect no digit in a test crop, although there should be at least one and at most five digits in each crop. \emph{Annotation-related-errors}: It is not uncommon to spot ground truth label errors in both the train and test datasets. The square crops are less tight than the rectangle ones, and they may include digits that should not be picked up as house numbers.

\subsection{Limitations of our method}

We discuss a few limitations of our method before concluding our paper. As most multiple object detection methods, our method requires a clustering stage, which could be error prone, to separate one instance from another. Our method may not be able to provide pixel-level object localization. Let an instance of the $\ell$th class is detected at location $(m, n)$. Still, we find that quantity
$\sum_{\rm R,\,G,\,B\, channels} \left| \frac{\partial \log p_{\ell, m, n}}{\partial \bold I}\right| $
does provide a vague image highlighting the region contributes the most to the detection of this instance. Strongly correlated instance labels might lead to less efficient learning. In the extreme case where at least one instance of each class is included in each training sample, nothing can be learned since one can simply set the model likelihood to $1$. Nevertheless, these limitations are the natural consequences due to our weak supervision settings, and are not unique to our method.    

\section{Conclusions}

We have proposed a novel method for a family of multiclass multiple instance learning where the labels only suggest that any instance of a class exists in a sample or not. We discuss its relationship to several existing techniques, and demonstrate its applications to multiple object detection and localization. With our method, weak labels and simple models are shown to be able to solve tough problems like the Google street view house number sequence recognition in reasonably realistic settings. 

\section*{Appendix A: Proof of Proposition 1} 

We start from (\ref{prb_in_beta}), and repetitively apply (\ref{beta_recursive}) to replace $\beta$'s with $\alpha$'s. This process is tedious, but could evenly prove the correctness of Proposition 1. Starting from the end of (\ref{prb_in_beta}) could make this process more manageable. By expanding the term $\beta_{ \ell_1, \ldots, \ell_{i-1}, \ell_{i+1}, \ldots, \ell_L, \phi}$ in (\ref{prb_in_beta}) with (\ref{beta_recursive}), we obtain 
\begin{align}\nonumber 
& {\rm Prb}(\{\ell_1, \ldots, \ell_L\}|\bold{P}) = \\ \nonumber 
&\qquad  \alpha_{\ell_1, \ldots, \ell_L, \phi} + (L-1)\beta_{\phi} + (L-2) \sum_{i=1}^L (  \beta_{\ell_i} + \beta_{\ell_i, \phi}) \\  \nonumber 
& \qquad + (L-3)\sum_{i=1}^L\sum_{j=i+1}^L ( \beta_{ \ell_i, \ell_j} + \beta_{ \ell_i, \ell_j, \phi}) - \ldots \\ \nonumber 
& \qquad +  \sum_{i=1}^L\sum_{j=i+1}^L \beta_{ \ell_1, \ldots, \ell_{i-1}, \ell_{i+1}, \ldots, \ell_{j-1}, \ell_{j+1}, \ldots, \ell_L, \phi} \\ \label{prb_in_beta2}
& \qquad - \sum_{i=1}^L  \alpha_{ \ell_1, \ldots, \ell_{i-1}, \ell_{i+1}, \ldots, \ell_L, \phi}
\end{align}
Next, we expand all the terms like  
$$\beta_{ \ell_1, \ldots, \ell_{i-1}, \ell_{i+1}, \ldots, \ell_{j-1}, \ell_{j+1}, \ldots, \ell_L, \phi}$$
in (\ref{prb_in_beta2}) using (\ref{beta_recursive}) to have
\begin{align*}\nonumber 
& {\rm Prb}(\{\ell_1, \ldots, \ell_L\}|\bold{P}) = \\ \nonumber 
&\qquad  \alpha_{\ell_1, \ldots, \ell_L, \phi} + [(L-1) - 0.5L(L-1)]\beta_{\phi} \\
& \qquad + [ (L-2) - (0.5L-1)(L-1) ]  \sum_{i=1}^L (  \beta_{\ell_i} + \beta_{\ell_i, \phi}) \\
& \qquad + \cdots  \\  \nonumber 
& \qquad +  \sum_{i=1}^L\sum_{j=i+1}^L \alpha_{ \ell_1, \ldots, \ell_{i-1}, \ell_{i+1}, \ldots, \ell_{j-1}, \ell_{j+1}, \ldots, \ell_L, \phi} \\
& \qquad - \sum_{i=1}^L  \alpha_{ \ell_1, \ldots, \ell_{i-1}, \ell_{i+1}, \ldots, \ell_L, \phi}
\end{align*}
We continue this process until all $\beta$'s are replaced with $\alpha$'s. Finally, the coefficient before $\alpha_{\phi}$ will be
\[ -1 + {L\choose 1} - {L\choose 2} + \ldots  - (-1)^{L-1} {L\choose L-1}  \] 
which is just $(-1)^L - [1 + (-1)]^L = (-1)^L$, where 
$${n \choose k} = \frac{n!}{(n-k)!k!}, \quad 1\le k\le n$$ 
denotes binomial coefficient. Similarly, the coefficient before terms like $\alpha_{\ell_i,\phi}, \alpha_{\ell_i,\ell_j,\phi}, \ldots$ will be 
\begin{align*}
(-1)^{L-1} - [1 + (-1)]^{L-1} & = (-1)^{L-1} \\
(-1)^{L-2} - [1 + (-1)]^{L-2} & = (-1)^{L-2} \\
& \vdots
\end{align*}
This finishes the proof of Proposition 1.

\bibliography{example_paper}
\bibliographystyle{icml2019}

\newpage
\section*{Supplementary Material A: Performance of traditional MIL}

We test the traditional MIL on the extended MNIST task. Please check Section 5.2 and code 

\url{https://github.com/lixilinx/MCMIL/blob/master/TraditionalMIL_ExtendedMNISTExperiment.py}

for details. Assume $\mathbb{L}$ and $\bold P$ are the bag label and probability tensor of an image $\bold I$, respectively. The cost function is 
\begin{equation}\label{traditional_mil_cost}
- \frac{1}{|\mathbb{L}|}\sum_{\ell \in \mathbb{L}} \left( \log \max_{m,n}p_{\ell, m, n} \right)
\end{equation}
For a batch of pairs $(\bold I, \mathbb{L})$, we minimize the batch averaged cost. It is crucial to normalize $\bold P$ as
\[ \sum_{\ell=1}^{C+1}\sum_{m=1}^M\sum_{n=1}^N p_{\ell,m,n} = 1 \] 
to make (\ref{traditional_mil_cost}) a valid cost for multiclass MIL. 
The normalization in (\ref{normalized_P}) does not work here since (\ref{traditional_mil_cost}) is not the true negative logarithm likelihood. With these settings, test classification error rates of six runs starting from random initial guesses are $8.9\%, 5.4\%, 8.4\%, 7.1\%, 10.1\%$ and $8.1\%$. MIL performs significantly worse than our method and strongly supervised baseline (our typical test error rates are below $0.4\%$).   

One may wonder that why MIL is shown to be successful on many other applications, but performs not so well at our tasks. One reason is that MIL can take many forms. Some are particularly suitable for specifics tasks and performs well, but not good at all tasks. Another reason is that we are comparing MIL with strongly supervised oracles, while many papers compare MIL with similar weak supervision methods. We also find that the performance gap between MIL and strongly supervised oracles is huge in the results reported in \cite{Pathak15}. One more reason might be the model optimization difficulty of our settings. Note that the $\max$ operation in (\ref{traditional_mil_cost}) is non-differentiable, but still convex. When (\ref{traditional_mil_cost}) is used along with convex models, e.g., support vector machine (SVM), the cost function is  convex with respect to model parameters, and relatively easy to be optimized. Actually, most MIL uses convex models, e.g., SVM \cite{Carbonneau18, Zhang14}. However, these shallow convex models do not perform well for complicated tasks like multiple object detection and localization. The cost function in (\ref{traditional_mil_cost}) becomes hard to optimize when highly non-convex models, e.g., deep neural networks, are used. Thus, when the traditional MIL is used for training non-convex models, pretraining might be essential for good performance \cite{Pathak15}. Unfortunately, finding a good initial guess is not always trivial. Region proposal may be required as well when applying MIL to object detection.  Our method is derived from a more general and principled approach, and seems do not suffer from these issues in our experiments.      

\end{document}